\documentclass[journal,10pt]{IEEEtran}

\usepackage{cite}
\usepackage{graphicx}
\usepackage[caption=false,font=footnotesize]{subfig}
\usepackage{booktabs}
\usepackage{multirow}
\usepackage{tabularx}
\usepackage{array}
\usepackage{xcolor}
\usepackage{url}

\graphicspath{{figures/}}
\newcolumntype{Y}{>{\raggedright\arraybackslash}X}
\definecolor{CoverFull}{HTML}{2A9D8F}
\definecolor{CoverPart}{HTML}{E9A03A}
\definecolor{CoverOpen}{HTML}{B8C0C8}

\begin{document}
\bstctlcite{IEEEtranBSTCTL}

\title{Towards CSI-Native Foundation Models:\\ A Channel-Adaptive Roadmap for 6G}

\author{Chenyu Zhang, Xinchen Lyu, Chenshan Ren, Shuhan Liu, and Qimei Cui%
    \thanks{C. Zhang, X. Lyu, and Q. Cui are with the National Engineering Research Center for Mobile Network Technologies, Beijing University of Posts and Telecommunications, Beijing 100876, China (e-mails: \{zhangchenyu2024, lvxinchen, cuiqimei\}@bupt.edu.cn).}
    \thanks{C. Ren is with the Key Laboratory of Ethnic Language Intelligent Analysis and Security Governance of MOE, Minzu University of China, Beijing 100081, China (e-mail: renchenshan06@163.com).}
    \thanks{S. Liu is with China Telecom Corporation Limited Gansu Branch, Gansu 730000, China (e-mail: liush20@chinatelecom.cn).}
}

\maketitle

\begin{abstract}
Wireless foundation models offer a path toward reusable channel state information (CSI) intelligence for sixth-generation (6G) systems. However, existing generic-backbone adaptation and CSI pretraining methods often treat CSI as task tensors rather than propagation-conditioned channel responses, thereby failing to capture the intrinsic time-frequency-spatial geometry of wireless environments. This paper presents a channel-adaptive roadmap toward CSI-native foundation models, proposing a unified framework that aligns pretraining, positional modeling, and attention control with three channel requirements: scale-aware heterogeneous exposure, physical time-frequency-antenna coordinates, and correlation-bounded token interaction. Extensive experiments demonstrate the superiority of the proposed framework across three dimensions: \textbf{zero-shot generalization} (reducing NMSE by $>4$~dB across spatial-temporal-frequency tasks), \textbf{scale extrapolation} (yielding up to a $5.4$~dB gain under $8\times$ unseen antenna scaling), and \textbf{inference efficiency} (accelerating mobility-aware processing by up to $18.8\%$).  A system-level evaluation with Sionna SYS further shows that the proposed framework uses only 7.01\% of dense-pilot overhead, reaches $-18.64$~dB average NMSE, and improves average net spectral efficiency by 36.6\% over dense LMMSE and 15.5\% over WiFo, indicating that CSI-native representation learning can support pilot-efficient radio access.
\end{abstract}


\section{Introduction}

The transition toward sixth-generation (6G) networks is marked by a fundamental shift in how artificial intelligence (AI) is integrated into the physical layer~\cite{farhadi20256g,cui2025overviewai6g}. The community is moving away from isolated, task-specific predictors toward foundation models (FMs) capable of reusable, generalized intelligence~\cite{zou2026llm6g}. At the heart of this revolution lies channel state information (CSI), the fundamental digital footprint of wireless propagation. If the rich, multi-dimensional observations of CSI can be captured by a unified foundation model, 6G systems could achieve unprecedented generalization across diverse channels, deployment scenarios, and downstream tasks such as channel prediction, beam management, and adaptive radio access~\cite{li2023aicsi}.

As illustrated in Fig.~\ref{fig:intro_wireless_fm_csi_gap}, recent explorations on wireless foundation models can be broadly categorized into two technical routes: 1) \emph{Generic Backbone Adaptation}, which transfers pretrained language or vision models to wireless tasks~\cite{liu2024llm4cp,guo2024lvm4csi,sun2025llm4pg}; and 2) \emph{CSI Model Pretraining}, which learns representations directly from wireless datasets~\cite{liu2025wifo,alikhani2024lwm,bian2026airfmdda}. While these routes represent crucial first steps, they share a fundamental limitation by reducing CSI to generic, task-specific tensors, thereby overlooking that the intrinsic features of CSI transcend basic data modalities. Unlike the semantic continuity of language or the local visual regularities of images, CSI is a highly dynamic, propagation-conditioned response shaped by complex scattering environments, mobility patterns, and multi-antenna geometries. When foundation models process CSI without accounting for this profound modality gap, they would suffer from the structural mismatch. To this end, we must design CSI-native foundation models that understand and embed these unique, channel-adaptive features directly into the lifecycle of model pretraining:

\begin{figure*}[t]
  \centering
  \includegraphics[width=\textwidth]{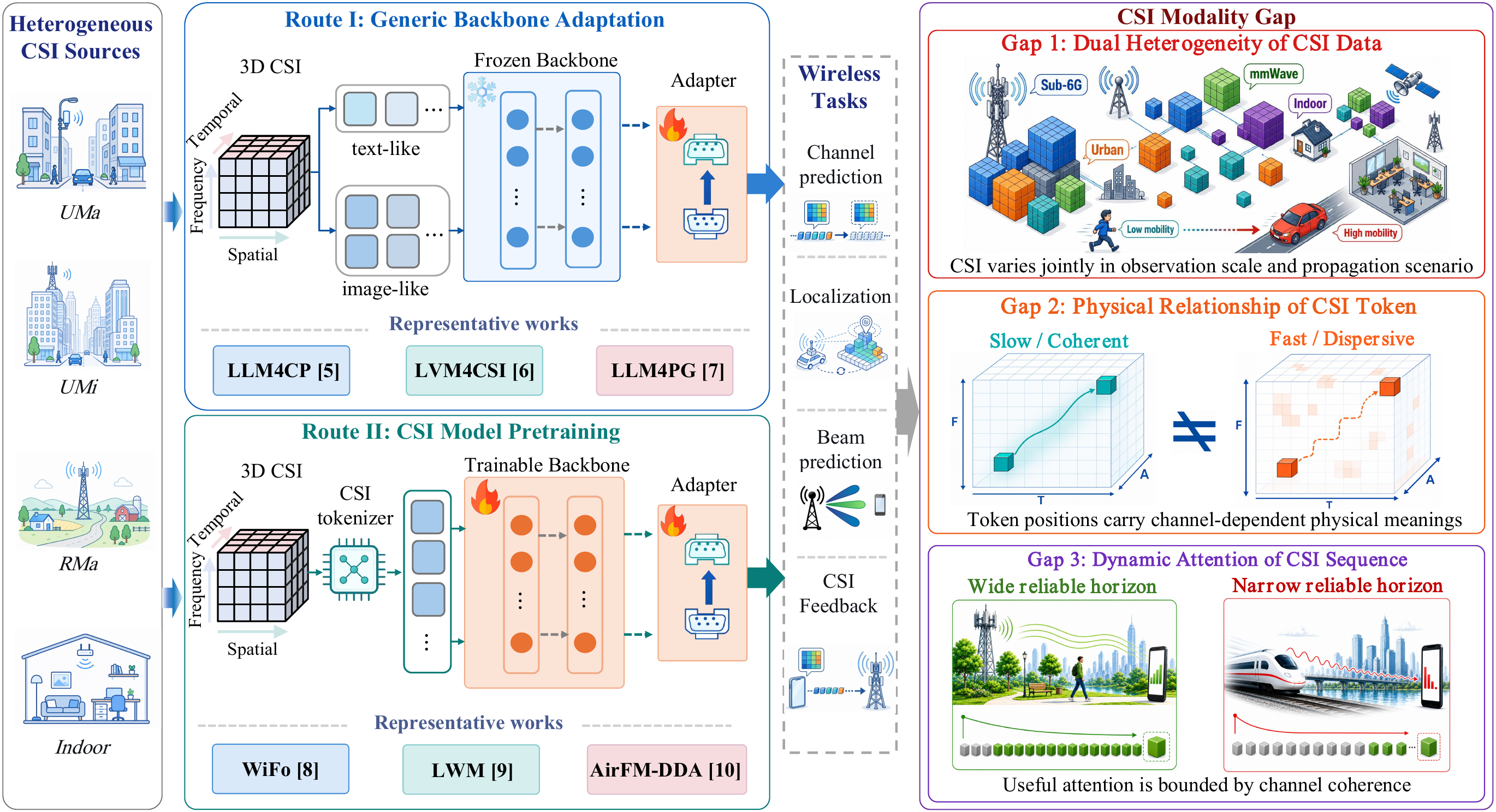}
  \caption{Two routes toward wireless foundation models and the CSI modality gap. Generic-backbone adaptation and CSI model pretraining improve wireless model reuse, but both must address the propagation-conditioned scale, coordinate, and correlation structure of CSI.}
  \label{fig:intro_wireless_fm_csi_gap}
\end{figure*}

\textit{1) Dual Heterogeneity of CSI Data: How to decouple hardware-induced scale variations from environment-driven propagation scenarios during large-scale pretraining?} Unlike standard image resolutions, CSI dimensions (time, frequency, antennas) are strictly tied to hardware configurations (scale), while its statistical distributions are dictated by physical environments like indoor, urban macro, or high-speed rail (scenario). Blindly mixing heterogeneous CSI in a generic training pipeline destabilizes learning process.

\textit{2) Physical Relationship of CSI Tokens: How to translate static token indices into dynamic physical offsets that capture time-frequency-space channel correlations?} In a standard transformer, a token's position is a sequential index. In CSI, a coordinate is a physical offset where temporal, frequency, and antenna distances reflect mobility, delay spread, and array geometry. Such physical correlation dynamically stretches or shrinks based on the environment, necessitating channel-adaptive positional encodings for CSI tokens.

\textit{3) Dynamic Attention of CSI Sequences: How to bound attention windows by the channel's physical coherence limits to avoid aggregating stale, uncorrelated noise?} In natural language processing, expanding the context window generally provides more semantic background. In wireless channels, however, the ``reliable context'' is strictly bounded by physical coherence time and bandwidth. Attending to CSI outside this relational validity horizon introduces destructive interference rather than knowledge.

In this article, we present a comprehensive channel-adaptive roadmap for transitioning toward CSI-native foundation models, and introduce a unified framework that explicitly aligns the model pretraining lifecycle with wireless channel characteristics. Specifically, we detail three coordinated mechanisms: 

\textit{1) Heterogeneity-Aware Data Scheduling:} We decouple hardware-induced scale variations from environment-driven scenarios through scale-compatible yet scenario-diverse pretraining. This structured exposure prevents overfitting to a fixed CSI grid and improves zero-shot NMSE by 7.19~dB, 4.08~dB, and 5.27~dB on CSI reconstruction, temporal prediction, and frequency extrapolation, respectively.

\textit{2) Channel-Adaptive 3D Positional Encoding:} We design a structural module that converts CSI token indices into channel-aware time-frequency-antenna offsets. By adapting positional phases to channel coherence, the module enables the foundation model to interpret CSI geometry under unseen antenna, temporal, and frequency configurations, thereby improving extrapolation beyond the scales observed during pretraining.

\textit{3) Correlation-Aware Attention Control:} We develop a relational mechanism that dynamically bounds context windows according to the physical coherence limits of the channel. Instead of attending uniformly to all tokens, the model selects reliable channel evidence and suppresses stale or weakly correlated interactions, reducing measured inference latency by up to 18.8\%.

System-level evaluations further verify that these CSI-native components effectively translate representation-level gains into pilot-efficient radio access. In a Sionna SYS uplink scenario, the proposed framework maintains an average NMSE of $-18.64$~dB while operating at a mere 7.01\% dense-pilot overhead. Moreover, it achieves average net spectral efficiency improvements of 36.6\% and 15.5\% over the dense LMMSE baseline and WiFo~\cite{liu2025wifo}, respectively. 

\section{The Wireless Modality Gap: From Generic Backbones to CSI-Native Pretraining}

To build a truly generalizable wireless foundation model, it is necessary to first understand why simply porting successful AI architectures from other domains falls short. The transition toward CSI-native pretraining is driven by a modality gap between generic data structures and the propagation-conditioned nature of wireless channels.

\begin{figure*}[t]
  \centering
  \includegraphics[width=\textwidth]{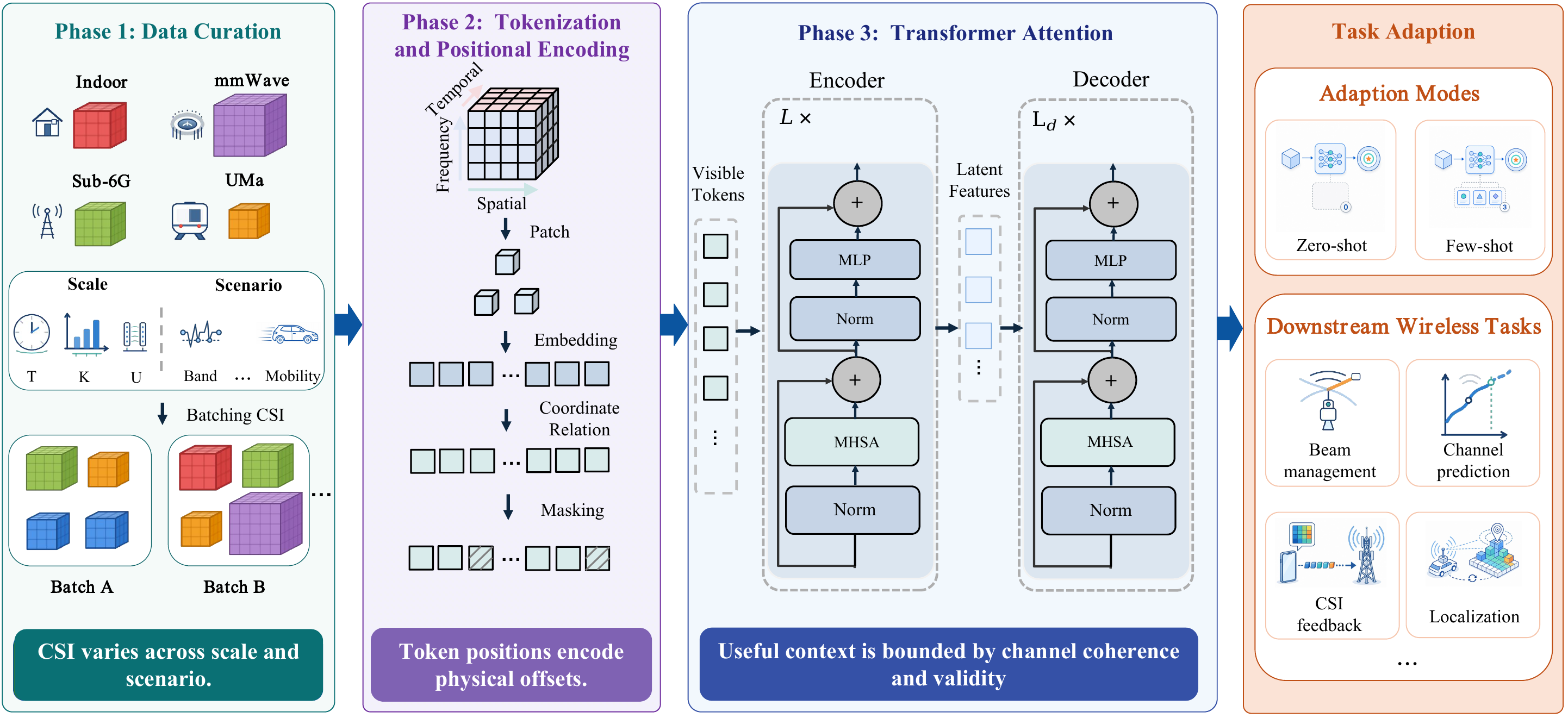}
  \caption{Training lifecycle roadmap for channel-adaptive CSI representation. The roadmap maps the CSI modality gap to three design layers: scale-aware data exposure, physical time-frequency-antenna geometry, and correlation-bounded token interaction.}
  \label{fig:roadmap_csi_native_lifecycle}
\end{figure*}

\subsection{The Current Landscape of Wireless Foundation Models}

Recent efforts to bring foundation models to the physical layer have largely followed two pioneering routes. The first is \emph{Generic Backbone Adaptation}, which leverages the progress of natural language processing (NLP) and computer vision (CV) by transferring pretrained large language models (LLMs) or large vision models (LVMs) to wireless tasks. Representative works such as LLM4CP, LVM4CSI, and LLM4PG achieve this through modality conversion, or fine-tuning~\cite{liu2024llm4cp,guo2024lvm4csi,sun2025llm4pg}.

The second route is \emph{CSI Model Pretraining}, which bypasses external backbones and trains models directly on large-scale wireless datasets. WiFo pretrains on CSI data for channel prediction~\cite{liu2025wifo}; LWM learns task-agnostic contextual channel representations~\cite{alikhani2024lwm}; AirFM-DDA builds an air-interface foundation model in the delay-doppler-angle domain for AI-native 6G~\cite{bian2026airfmdda}; and MCM unifies CSI feedback, prediction, and estimation through masked channel modeling~\cite{guo2026scalablemcm}. While both routes represent crucial steps beyond isolated, task-specific predictors, they still expose a common bottleneck: CSI is often reduced to generic task tensors, making it difficult for the model to preserve the channel evidence, coordinate semantics, and reliability constraints needed for robust generalization.

\subsection{Why CSI Transcends Basic Modalities}

The core limitation of generic backbone adaptation is that CSI does not follow the semantic continuity of human language or the local visual regularities of images. Instead, CSI is a highly dynamic, propagation-conditioned response governed by complex scattering environments, mobility patterns, and multi-antenna geometries. When an LLM or LVM processes CSI, it must expend immense representational capacity simply compensating for this modality mismatch before it can begin to learn reusable channel behavior. Furthermore, generic large models are computationally heavy, making them impractical for the latency- and energy-constrained edge deployments required in 6G Radio Access Networks (RAN)~\cite{zou2026llm6g}.

To overcome this inherent CSI modality gap, models must translate raw physical evidence into generalizable representations while preserving essential properties like physical scale, spatial coordinates, and correlation structures. This necessitates a paradigm shift toward CSI-native pretraining, where representations are learned directly from wireless signals. However, simply feeding CSI into a standard transformer architecture is insufficient. If the training process does not explicitly account for the physical meaning and reliability of CSI evidence, the model will merely memorize incompatible channel tensors rather than learning generalizable wireless physics. To bridge this gap, the entire AI lifecycle must be redesigned to be channel-adaptive.


\subsection{The Channel-Adaptive AI Lifecycle}

Designing a CSI-native foundation model requires rethinking the standard transformer pipeline across three layers, as conceptualized in the training lifecycle roadmap of Fig.~\ref{fig:roadmap_csi_native_lifecycle}. These layers map the wireless modality gap into three concrete barriers that must be addressed before CSI can become a reusable foundation-model modality.

\emph{Phase 1: Data Curation (The Data Layer).} The first hurdle appears before training begins. CSI samples vary simultaneously in hardware scale, such as time length, bandwidth, and antenna count, and in propagation scenario, such as carrier band, mobility, indoor/outdoor deployment, and scattering condition. A generic data pipeline that blindly shuffles these variations may force the model to reconcile structurally incompatible CSI grids within the same update, destabilizing representation learning and obscuring useful scenario diversity.

\emph{Phase 2: Tokenization and Positional Encoding (The Structural Layer).} Transformers require positional information because self-attention is permutation equivariant~\cite{vaswani2017attention}. Yet a CSI token position is not a generic sequence index; it is a physical offset in the time-frequency-antenna domain. Because the correlation between CSI tokens can stretch or shrink with mobility, delay spread, carrier frequency, and array geometry, static or flattened positional encodings fail to capture the true channel geometry.

\emph{Phase 3: Transformer Attention (The Relational Layer).} In language modeling, expanding the attention window generally provides more semantic context. In wireless channels, however, reliable context is bounded by physical coherence time, coherence bandwidth, and spatial correlation. Standard global attention may aggregate stale or weakly correlated tokens outside this validity horizon, degrading both accuracy and inference efficiency. Addressing the three lifecycle barriers therefore requires a unified framework that embeds wireless-channel awareness into data scheduling, positional geometry, and attention control.

\section{A CSI-native Foundation Model: Channel-Adaptive Representation Learning}

This section presents a unified framework for CSI-native foundation models through channel-adaptive representation learning. The core design principle is to embed wireless-channel awareness into the points where CSI is exposed, indexed, and correlated, rather than relying on model scale alone to recover channel structure from flattened tensors.

\subsection{Heterogeneity-Aware Data Scheduling}

Existing pretraining studies predominantly treat CSI heterogeneity as scenario variations, such as shifts in frequency bands~\cite{liu2024llm4cp,alikhani2024lwm}. WiFo~\cite{liu2025wifo} and WiFo-2~\cite{liu2025wifo2} further expand on this concept by recognizing that CSI also exhibits scale heterogeneity, including fluctuations in the number of subcarriers, antennas, and temporal samples. Accordingly, they employ multi-dataset pre-training with ViT backbones to derive more generalized channel representations. However, these methods do not explicitly separate hardware-induced scale variations from environment-driven propagation changes, thereby conflating two distinct sources of CSI heterogeneity and making it difficult to evaluate their respective impacts on model pretraining.

\begin{figure}[!t]
  \centering
  \subfloat[Module A: Heterogeneity-aware data scheduling]{
    \includegraphics[width=\columnwidth]{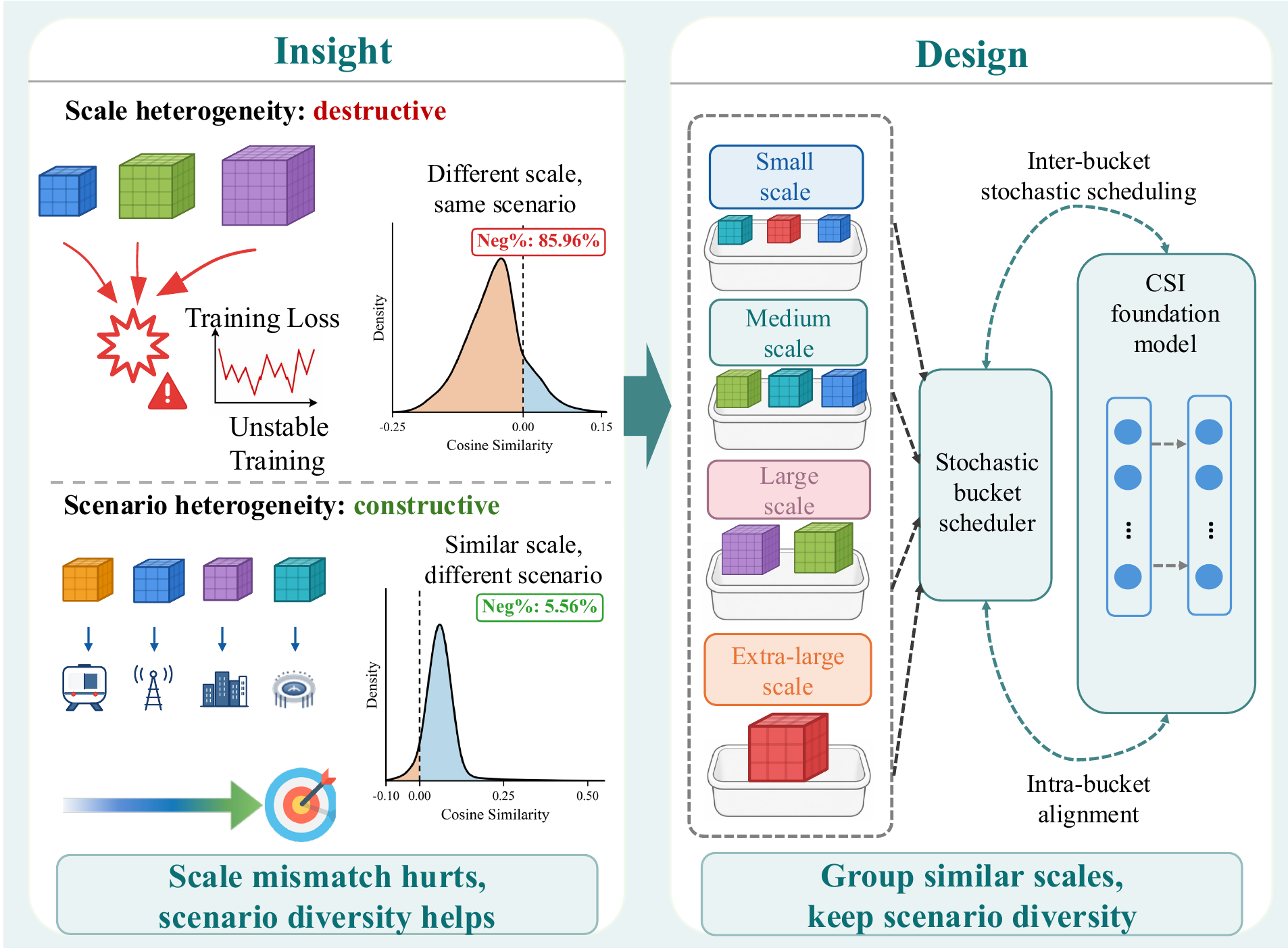}
    \label{fig:framework_data_scheduling}
  }\\[-0.2ex]
  \subfloat[Module B: Channel-adaptive 3D positional encoding]{
    \includegraphics[width=\columnwidth]{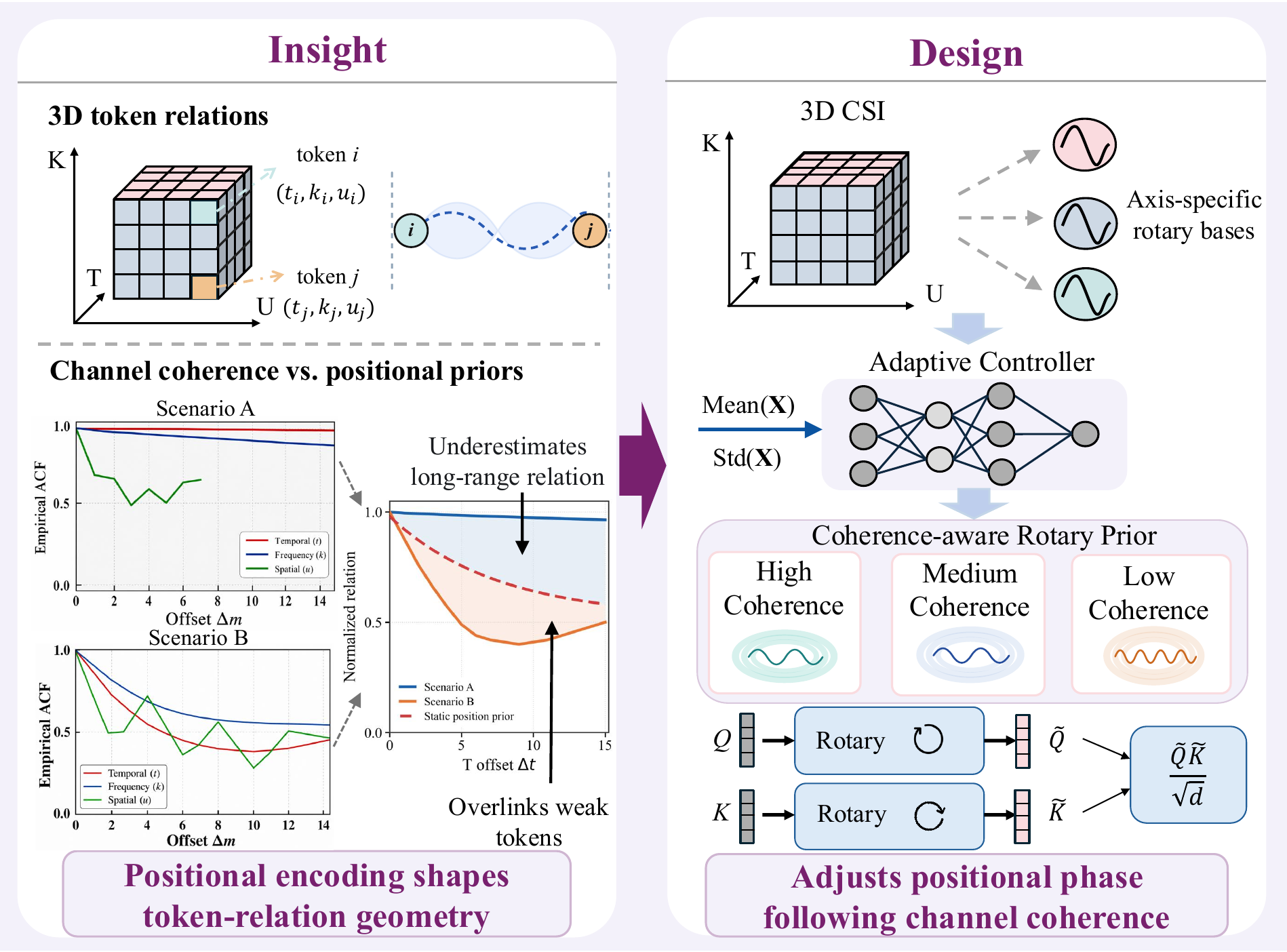}
    \label{fig:framework_adaptive_3d_rope}
  }\\[-0.2ex]
  \subfloat[Module C: Correlation-aware attention control]{
    \includegraphics[width=\columnwidth]{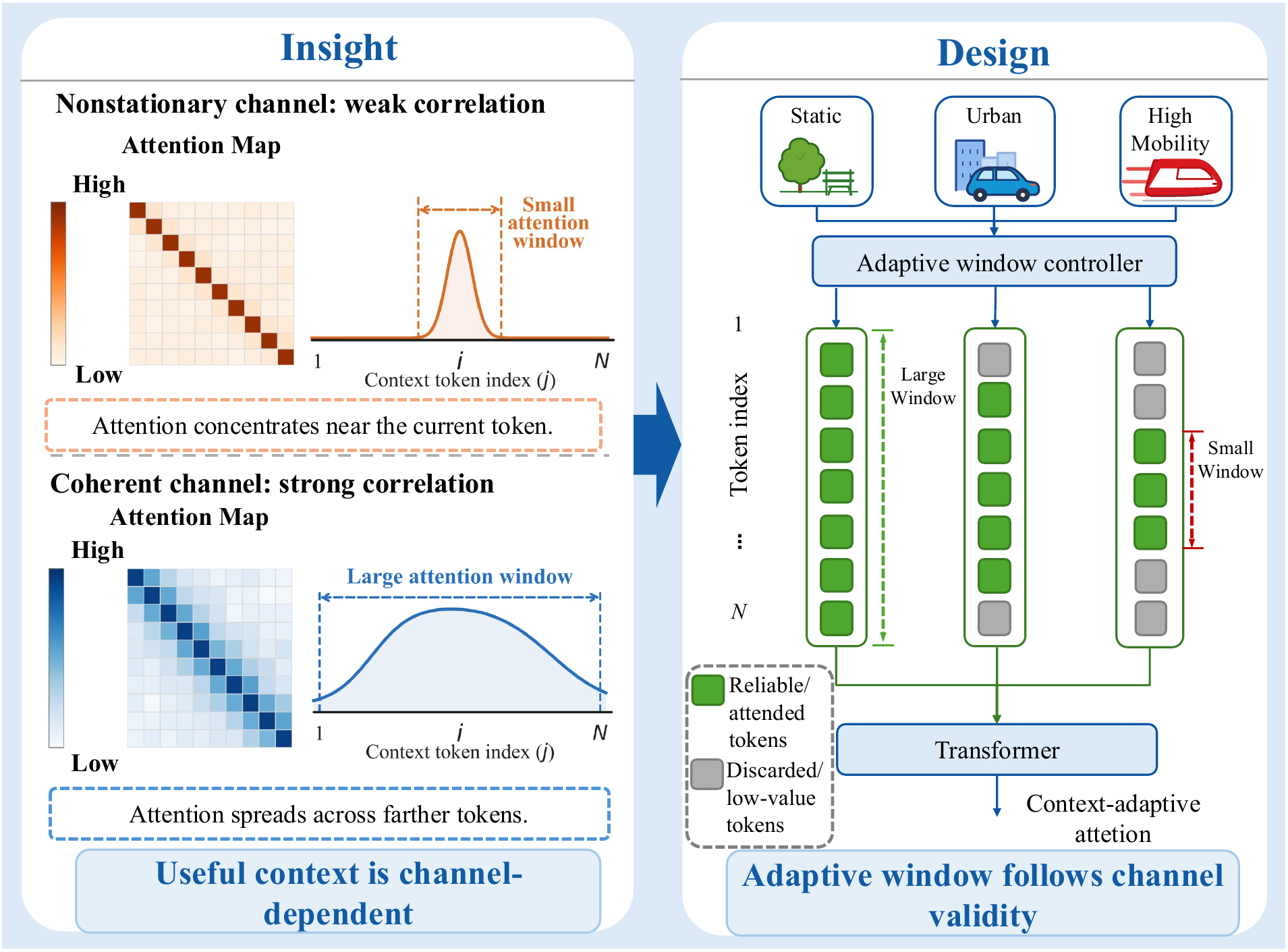}
    \label{fig:framework_correlation_attention}
  }
  \caption{Three modules of the proposed channel-adaptive CSI foundation modeling framework and their mapping to the validation cases.}
  \label{fig:framework_overview}
\end{figure}

\emph{Insight:} As shown in Fig.~\ref{fig:framework_overview}(a), CSI heterogeneity stems from two distinct dimensions: hardware-induced \emph{scales} (e.g., temporal duration, subcarrier grid, and antenna count) and environment-driven \emph{scenarios} (e.g., carrier band, mobility, and scattering richness). Blindly mixing these variations in a standard training pipeline destabilizes the learning process. Scale-mismatched samples force the model to reconcile structurally incompatible tensors within a single gradient update, causing severe representation conflicts. Conversely, scenario diversity acts as a constructive regularizer that enriches channel representations, provided the structural scale is controlled. Therefore, our core design is to decouple scale variations from propagation scenarios, transforming uncontrolled dataset mixing into structured, scale-consistent CSI exposure~\cite{zhang2026hetercsi}.

\emph{Design:} The scheduler preserves scenario diversity while constraining each update to scale-compatible CSI observations through three steps: 1) \textit{Scale Descriptor Construction}: Each CSI sample is partitioned into uniform 3D patches over the time-frequency-antenna domains, and the resulting token count forms a compact scale descriptor. 2) \textit{Bucket-Constrained Batching}: Samples are grouped into percentile buckets by this descriptor, and each mini-batch is drawn from a single bucket so that every optimization step operates on compatible sequence supports. 3) \textit{Scenario-Preserving Shuffling}: Bucket boundaries are preserved during batching, while buckets are stochastically shuffled across epochs and distributed across training ranks to retain propagation diversity. Dynamic token masks then support scale-compatible gradient updates without discarding scenario-rich pretraining exposure.

\subsection{Channel-Adaptive 3D Positional Encoding}

Many CSI foundation models inherit one-dimensional absolute positional embeddings from LLMs, assigning sequential indices to flattened CSI patch tokens and thereby collapsing the native spatial, temporal, and frequency channel grid into a 1D sequence~\cite{vaswani2017attention}. WiFo~\cite{liu2025wifo} and WiFo-2~\cite{liu2025wifo2} retain original domain token structures with static 3D absolute positional encoding, while AirFM-DDA~\cite{bian2026airfmdda} introduces positional encoding aware of frame structures in the delay-doppler-angle domain. These designs better preserve location information, yet their static or grid-tied priors still cannot jointly provide explicit relative decay, preserve decoupled dimensions in 3D space, and adapt positional interactions to fluctuating channel coherence.

\emph{Insight:} Fig.~\ref{fig:framework_overview}(b) shows that CSI token relations are governed by multi-dimensional coherence across three physical axes rather than a generic sequential order. Specifically, temporal, frequency, and antenna offsets respectively reflect doppler, delay, and angular coherence. The same coordinate offset may imply strong or weak correlation under different scenarios, because mobility, carrier frequency, bandwidth, and scattering richness reshape the channel coherence profile. Such environment-sensitive dependency is the structural prior that positional encoding should provide at the representation level. A suitable design should preserve 3D CSI coordinates and adapt relative-distance priors to sample-specific channel coherence~\cite{zhang2026adaptive3drope}.

\emph{Design:} The module builds a channel-adaptive 3D rotary prior through three steps: 1) \textit{Axis-Specific Rotary Prior}: A learnable frequency bank is defined separately for the time, frequency, and antenna axes, forming a pretrained rotary prior over 3D CSI coordinates. 2) \textit{Sample-Wise Modulation}: A lightweight controller summarizes token embeddings by their mean and standard deviation, then predicts scale and bias factors that modulate the base frequency bank. 3) \textit{Masked Phase Application}: The modulated phases rotate query and key vectors before self-attention, with phases gathered only for CSI patches retained after masking.

\subsection{Correlation-Aware Attention Control}

Most existing CSI foundation models apply global self-attention to extract context from the full CSI grid~\cite{liu2025wifo,liu2025wifo2}. Although advances like AirFM-DDA~\cite{bian2026airfmdda} introduces window-based attention to exploit clustered multipath structures in the DDA domain, but its receptive fields remain predefined and content-agnostic. They do not explicitly adapt to the actual physical coherence boundaries of individual CSI samples. Lacking dynamic, correlation-aware regulation, such models are prone to incorporating stale temporal data, faded frequency signals, and misleading spatial interference into the attention formulation.

\emph{Insight:} As illustrated in Fig.~\ref{fig:framework_overview}(c), channel correlation reshapes the reliable evidence available to self-attention. After CSI tokens are embedded with coordinate-aware priors, the model must still decide which context tokens should participate in the current prediction. In a weakly correlated channel, useful evidence is concentrated near the query token, requiring a narrow and selective attention map; in a strongly correlated channel, a wider window can preserve valid long-range interactions. Correlation-aware attention therefore acts as a reliability filter, suppressing stale or weakly informative tokens while avoiding unnecessary global aggregation.

\emph{Design:} The module restricts attention to channel-supported token neighborhoods through three steps: 1) \textit{Correlation Context Construction}: Valid CSI tokens are mapped to the 3D time-frequency-antenna grid to estimate normalized axis-wise lag correlations. The maximum reliable lag forms the base window, while unestimated axes default to the valid global span. 2) \textit{Adaptive Window Regulation}: A lightweight controller combines the base window, compact physical indicators, and token/power statistics to produce a discrete hard window and a continuous soft window within predefined bounds. 3) \textit{Attention Execution}: The hard window defines the admissible 3D neighbor set, whereas the soft window applies a distance-dependent bias to suppress weakly correlated interactions. This confines token aggregation to the physically coherent channel context and reduces unnecessary computation.

\section{Validation Cases}

This section evaluates the proposed framework from two perspectives. Module-level validation isolates the effects of the three CSI-native components on zero-shot transfer, scale extrapolation, and efficient inference. System-level validation integrates the three components into a unified framework and examines whether the resulting CSI representation can support pilot-efficient radio access after lightweight adaptation to the target Sionna SYS scenario.

\emph{Module-Level Setup and Benchmarks:} The module-level evaluation utilizes QuaDRiGa CSI generated under 3GPP TR 38.901 assumptions for pretraining and testing across diverse deployment scenarios, carrier bands, mobility regimes, and antenna configurations. The benchmarks are categorized by evaluation target: 1) full-shot task-specific small models and WiFo~\cite{liu2025wifo} for zero-shot transfer; 2) conventional positional encodings (1D-APE, 3D-APE, and fixed 3D-RoPE) for scale extrapolation; and 3) global attention for measured inference-time comparisons.


\emph{System-Level Setup and Benchmarks:} We deploy a Sionna SYS uplink topology (3GPP TR 38.901 UMa) that integrates proportional-fair scheduling, link adaptation, PHY abstraction, and HARQ feedback. The network parameters include: 1) full-buffer traffic for $\{4,8,12,16\}$ UEs; 2) a 3.5~GHz carrier with 24 PRBs, 14 OFDM symbols, and 1000 slots; and 3) 4 BS receive antennas. The proposed framework is evaluated against dense LMMSE ($0.5\times0.5$ DMRS), an AI-free sparse baseline ($0.125\times0.25$ DMRS), and WiFo~\cite{liu2025wifo}. Our method and WiFo further reduces average overhead by performing pilot-free temporal prediction in slots where historical CSI context remains reliable.

\subsection{Module-Level Validation Cases}

\begin{figure}[!t]
  \centering
  \subfloat[Heterogeneity-aware pretraining: zero-shot NMSE gains across CSI reconstruction, temporal prediction, and frequency extrapolation.]{
    \includegraphics[width=0.96\columnwidth]{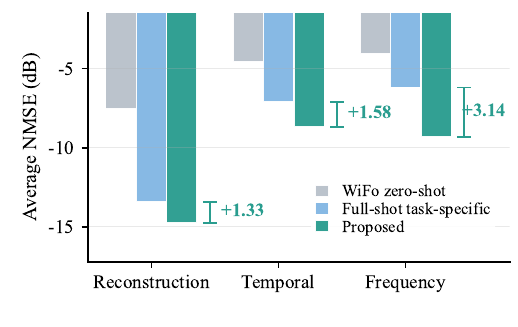}
    \label{fig:validation_heterogeneity_generalization}
  }\\[-0.2ex]
  \subfloat[Channel-adaptive 3D positional encoding: NMSE gains over positional baselines across antenna settings.]{
    \includegraphics[width=0.96\columnwidth]{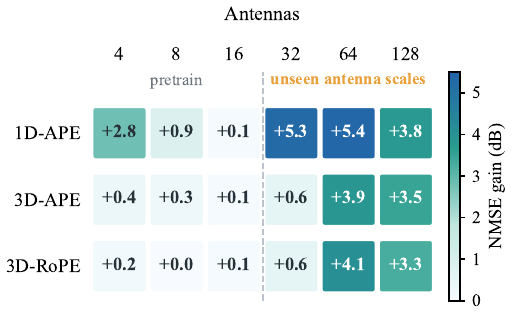}
    \label{fig:validation_adaptive_rope_extrapolation}
  }\\
  \subfloat[Correlation-aware attention control: batch inference-time reduction under medium/high mobility.]{
    \includegraphics[width=0.96\columnwidth]{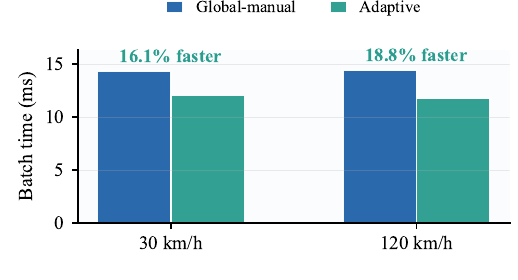}
    \label{fig:validation_correlation_attention_latency}
  }
  \caption{Validation results for the three channel-adaptive framework modules.}
  \label{fig:validation_module_results}
\end{figure}

\subsubsection{Case 1: Heterogeneity-Enhanced Generalization}

Fig.~\ref{fig:validation_module_results}(a) evaluates whether heterogeneity-aware pretraining learns transferable CSI representations across CSI reconstruction, temporal-domain prediction, and frequency-domain prediction. Compared with the WiFo zero-shot baseline~\cite{liu2025wifo}, the proposed framework reduces average NMSE by 7.19~dB, 4.08~dB, and 5.27~dB on the three tasks, respectively, and further outperforms full-shot task-specific small models by 1.33~dB, 1.58~dB, and 3.14~dB. These results show that controlled heterogeneous exposure preserves scenario diversity while avoiding scale-incompatible updates, enabling the pretrained model to learn reusable channel representations rather than fitting one task or scenario.

\subsubsection{Case 2: Position-Enabled Extrapolation}

Fig.~\ref{fig:validation_module_results}(b) reports NMSE gains of the proposed adaptive 3D positional encoding over representative positional baselines under different antenna numbers. As the antenna configuration changes, 1D positional encoding, 3D absolute positional encoding, and fixed 3D rotary encoding become less reliable because their positional priors do not adapt to changing channel geometry. The positive gains indicate that axis-decoupled 3D geometry with sample-wise modulation helps the model interpret channel-varying distance and extrapolate beyond the temporal, frequency, or antenna scales observed during pretraining.

\subsubsection{Case 3: Correlation-Aware Efficient Inference}

Fig.~\ref{fig:validation_module_results}(c) compares the batch inference time of a global baseline and a correlation-aware adaptive-window variant under the same heterogeneous data pipeline and positional encoding. At 30~km/h and 120~km/h, adaptive attention reduces average batch inference time by 16.1\% and 18.8\%, respectively. This confirms that when mobility shortens the temporal coherence horizon, correlation-aware attention can retain useful local evidence while removing token interactions that mainly add latency.

\subsection{System-Level Pilot-Efficient CSI Evaluation}

\begin{figure}[!t]
  \centering
  \subfloat[Pilot overhead--NMSE tradeoff.]{
    \includegraphics[width=0.82\columnwidth]{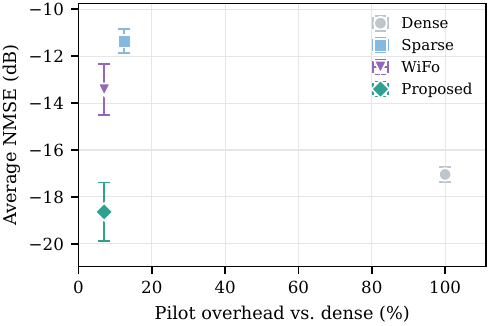}
    \label{fig:system_pilot_overhead_nmse}
  }\\[-0.2ex]
  \subfloat[Net spectral efficiency under different UE counts.]{
    \includegraphics[width=0.84\columnwidth]{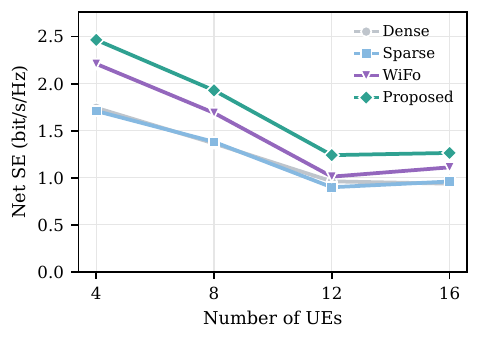}
    \label{fig:system_net_spectral_efficiency}
  }
  \caption{System-level pilot-efficient CSI evaluation with Sionna SYS.}
  \label{fig:system_pilot_efficiency}
\end{figure}

Fig.~\ref{fig:system_pilot_efficiency}(a) highlights the fundamental limitation of naive pilot reduction: the Sparse baseline degrades average NMSE to -11.36 dB. Even with foundation-model enhancement, WiFo~\cite{liu2025wifo} achieves only -13.43 dB at a pilot overhead of 7.01\% relative to the dense baseline. Conversely, the Proposed model effectively breaks this tradeoff, achieving an impressive -18.64 dB NMSE. It not only outperforms WiFo by 5.21 dB at the same overhead but also surpasses the dense LMMSE baseline by 1.60 dB. As shown in Fig.~\ref{fig:system_pilot_efficiency}(b), this superior CSI fidelity directly translates to enhanced net spectral efficiency, achieving 2.47, 1.93, 1.24, and 1.26 bit/s/Hz across the respective UE counts. Yielding average efficiency gains of 36.6\% over Dense and 15.5\% over WiFo, these results systematically prove that reducing pilot overhead is only viable when the underlying CSI adapter provides the rigorous reconstruction accuracy required to offset estimation-induced SINR degradation.

\section{Discussion and Future Directions}

The proposed framework represents an initial step toward CSI-native foundation models for 6G, where wireless intelligence is built around channel structure rather than adapted from generic data modalities. Beyond the three channel-adaptive modules studied in this paper, a complete roadmap must also address how CSI is tokenized, pretrained, transferred, deployed, fused with other wireless modalities, and updated after deployment. This section discusses these open directions and outlines how CSI foundation models may evolve from offline pretraining prototypes into practical, adaptive components of future 6G radio access networks.

\textit{1) CSI-Native Tokenizer:} A deployable wireless foundation model first needs a tokenizer that converts heterogeneous CSI into tokens that are both efficient and physically meaningful. Fixed patches are easy to implement but may waste tokens in coherent regions and lose details in fast-varying channel regions. Future tokenizers should adapt token granularity along time, frequency, and antenna axes according to channel coherence, phase dynamics, and delay-Doppler-angular structure.

\textit{2) CSI Foundation Model Pretraining Objective:} The next challenge is to define pretraining tasks that teach reusable wireless channel representations rather than only local reconstruction. Masked CSI modeling, temporal prediction, frequency extrapolation, denoising, contrastive scenario alignment, and metadata-conditioned objectives may capture different channel properties. A strong objective should connect self-supervised learning with downstream uses such as beam management, feedback compression, sensing, and digital-twin calibration.

\textit{3) Sim-to-Real and Few-Shot Adaptation:} Large CSI models will likely be trained on abundant simulated or digital-twin data, but deployed on hardware with calibration errors, synchronization offsets, RF impairments, and measurement noise. Bridging this gap requires few-shot adaptation, domain calibration, uncertainty estimation, and simulator refinement. The goal is not to replace measurements with simulation, but to make simulated pretraining transferable to real cells, bands, and mobility profiles with minimal target-domain data.

\textit{4) Deployment-Aware CSI Large Models:} A CSI foundation model must eventually run under the latency, memory, and energy constraints of real UE and gNB pipelines. This makes compression, sparse attention, quantization, early exiting, hardware-aware batching, and split inference central design issues. The deployment target should influence the model from the beginning: a useful CSI large model should expose controllable accuracy-latency tradeoffs rather than only maximizing offline NMSE.

\textit{5) ISAC and Digital-Twin Multimodal Fusion:} Future 6G systems will not rely on CSI alone. Integrated sensing and communication (ISAC), radio maps, positioning traces, beam measurements, environment semantics, and digital-twin states can provide complementary context for channel-adaptive representation. A promising direction is multimodal wireless foundation modeling, where CSI tokens are aligned with sensing, geometry, and network-state tokens to support prediction, planning, and validation in a unified representation space.


\section{Conclusion}

This paper establishes that advancing toward CSI-native foundation models requires more than applying generic backbones over wireless tensors. Given that CSI fundamentally embeds hardware-dependent scales, physical time-frequency-antenna coordinates, and coherence-bounded reliability, mechanisms like data scheduling, positional modeling, and attention control are not peripheral implementation choices, but architectural prerequisites for reusable wireless intelligence. To this end, the proposed channel-adaptive framework instantiates this principle by integrating scale-aware heterogeneous exposure, adaptive 3D positional encoding, and correlation-bounded token interaction. Both module- and system-level evaluations confirm that this physical alignment yields profound gains in zero-shot generalization, scale extrapolation, inference efficiency, and pilot-efficient radio access. Ultimately, this paper demonstrates that the future of CSI foundation models lies not in directly transplanting generic foundation-model recipes, but in a channel-adaptive lifecycle where core AI components are strictly co-designed with wireless propagation physics.

\bibliographystyle{IEEEtran}
\bibliography{refs}

@IEEEtranBSTCTL{IEEEtranBSTCTL,
  CTLdash_repeated_names = "no"
}

@article{cui2025overviewai6g,
  author={Q. Cui and X. You and N. Wei and others},
  title={Overview of {AI} and communication for {6G} network: Fundamentals, challenges, and future research opportunities},
  journal={Science China Information Sciences},
  volume={68},
  number={7},
  pages={171301},
  year={2025},
  doi={10.1007/s11432-024-4337-1}
}

@article{li2023aicsi,
  author={Y. Li and Y. Hu and K. Min and others},
  title={Applying {AI} to {CSI} for High Efficiency Wireless Communication},
  journal={IEEE Wireless Communications},
  volume={30},
  number={1},
  pages={104--110},
  year={2023},
  doi={10.1109/MWC.005.2200245}
}

@article{guo2026scalablemcm,
  author={Guo, Jianhua and Deng, Zhongsheng and Qiao, Zhen and others},
  journal={IEEE Transactions on Communications}, 
  title={Scalable Pre-Trained Masked Channel Model of Wireless Communications}, 
  year={2026},
  volume={74},
  number={},
  pages={6197-6212},
  doi={10.1109/TCOMM.2026.3675420}}

@article{zou2026llm6g,
  author={H. Zou and Q. Zhao and S. Lasaulce and others},
  title={Large language models in {6G} from standard to on-device networks},
  journal={Nature Reviews Electrical Engineering},
  volume={3},
  pages={123--134},
  year={2026},
  doi={10.1038/s44287-025-00239-6}
}

@inproceedings{vaswani2017attention,
  author={A. Vaswani and N. Shazeer and N. Parmar and others},
  title={Attention is all you need},
  booktitle={Proc. Advances in Neural Information Processing Systems},
  year={2017}
}

@article{liu2024llm4cp,
  author={B. Liu and X. Liu and S. Gao and others},
  title={{LLM4CP}: Adapting large language models for channel prediction},
  journal={Journal of Communications and Information Networks},
  volume={9},
  number={2},
  pages={113--125},
  doi={10.23919/JCIN.2024.10582829},
  year={2024}
}

@article{liu2025wifo,
  title={{WiFo: Wireless foundation model for channel prediction}},
  author={B. Liu and S. Gao and X. Liu and others},
  journal={Science China Information Sciences},
  volume={68},
  number={6},
  pages={162302},
  year={2025},
  publisher={Springer}
}

@article{liu2025wifo2,
  author={B. Liu and X. Liu and S. Gao and others},
  title={{WiFo-2}: A generalist foundation model unifies heterogeneous wireless system design},
  journal={arXiv preprint arXiv:2511.22222},
  year={2025}
}

@article{guo2024lvm4csi,
  author={J. Guo and P. Jiang and C.-K. Wen and others},
  title={{LVM4CSI}: Enabling direct application of pre-trained large vision models for wireless channel tasks},
  journal={arXiv preprint arXiv:2507.05121},
  year={2025}
}

@article{alikhani2024lwm,
  author={S. Alikhani and G. Charan and A. Alkhateeb},
  title={Large wireless model ({LWM}): A foundation model for wireless channels},
  journal={arXiv preprint arXiv:2411.08872},
  year={2024}
}

@article{sun2025llm4pg,
  author={M. Sun and L. Bai and X. Cheng and others},
  title={{LLM4PG}: Adapting large language model for pathloss map generation via synesthesia of machines},
  journal={arXiv preprint arXiv:2511.02423},
  year={2025}
}

@article{bian2026airfmdda,
  title={{AirFM-DDA: Air-Interface Foundation Model in the Delay-Doppler-Angle Domain for AI-Native 6G}},
  author={Bian, Kejia and Tao, Meixia and Mo, Jianhua and others},
  journal={arXiv preprint arXiv:2605.00020},
  year={2026}
}

@article{zhang2026hetercsi,
  author={C. Zhang and X. Lyu and C. Ren and others},
  title={{HeterCSI}: Channel-adaptive heterogeneous {CSI} pretraining framework for generalized wireless foundation models},
  journal={arXiv preprint arXiv:2601.18200},
  year={2026}
}

@article{zhang2026adaptive3drope,
  author={C. Zhang and X. Lyu and C. Ren and others},
  title={Adaptive {3D-RoPE}: Physics-aligned rotary positional encoding for wireless foundation models},
  journal={arXiv preprint arXiv:2605.00968},
  year={2026}
}

@article{farhadi20256g,
  title={{6G} {AI}-Driven Air Interface---{Hexa-X-II} View},
  author={Farhadi, Hamed and Banerjee, Bitan and Berkvens, Rafael and others},
  journal={IEEE Communications Magazine},
  volume={63},
  number={10},
  pages={118--125},
  month=oct,
  year={2025},
  doi={10.1109/MCOM.001.2400394}
}

\end{document}